\documentclass{article}
\usepackage{amsmath,graphicx}
\usepackage{spconf}
\usepackage{cite}
\usepackage{array}
\usepackage{multirow}
\usepackage{balance}

\newcolumntype{x}[1]{%
>{\centering\hspace{0pt}}p{#1}}%

\graphicspath{{../figures/}}

\title{Atmospheric turbulence mitigation for sequences with moving objects using recursive image fusion}
\name{N. Anantrasirichai, Alin Achim, David Bull\thanks{This work was supported by EPSRC Platform Grant, Vision for the Future (EP/M000885/1).}}
\address{Visual Information Laboratory, University of Bristol, UK}

\begin{document}
%
\maketitle
\begin{abstract}

This paper describes a new method for mitigating the effects of atmospheric distortion on observed sequences that include large moving objects. In order to provide accurate detail from objects behind the distorting layer, we solve the space-variant distortion problem using recursive image fusion based on the Dual Tree Complex Wavelet Transform (DT-CWT). The moving objects are detected and tracked using the improved Gaussian mixture models (GMM) and Kalman filtering. New fusion rules are introduced which work on the magnitudes and angles of the DT-CWT coefficients independently to achieve a sharp image and to reduce atmospheric distortion, respectively.
The subjective results show that the proposed method achieves better video quality than other existing methods with competitive speed. 
\end{abstract}


\section{Introduction}
\label{sec:intro}

Atmospheric turbulence effects in acquired imagery make it extremely difficult to interpret the information behind the distorted layer typically formed by temperature variations or aerosols. This occurs when an object, e.g. the ground or the air itself, is hotter than the surrounding air. In such cases, the air is heated and begins to form horizontal layers. Increasing the temperature difference leads to faster and greater micro-scale changes in the air's refractive index. This effect is observed as a change in the interference pattern of the light refraction and causes the contents of the images and videos to appear shifted from their actual positions.  The main problem is that these movements are random, spatially and temporally varying perturbations, making a model-based solution difficult, particularly for sequences with large moving objects.

High-speed cameras can be used with a short exposure time to freeze moving objects so as to minimise distortions associated with motion blur. However, a geometric distortion, which is the result of anisoplanatic tip/tilt wavefront errors, will still be present. The atmospheric turbulence can be viewed as being quasi-periodic; therefore, averaging a number of frames yields a geometric improvement in the image, but it remains blurred by an unknown point spread function (PSF) of the same size as the pixel motions due to the turbulence. Experiments in \cite{Mao:nonrigid:2012} revealed that assuming a Gaussian blur kernel for non blind deconvolution or a Bayesian blind deconvolution cannot efficiently remove turbulent distortions and showed insignificantly different subjective results. Other techniques exploit a subset of the data by selecting the best quality in the temporal direction.  
However, it is almost impossible to discard the regions that include moving objects, whilst still maintaining smooth motion in videos. Most methods detect long-distance target objects that are at sufficiently low fidelity to exhibit little or no detail, instead appearing as blurred silhouettes or blobs \cite{ Oreifej:Simultaneous:2013, Chen:detecting:2014}.
A few methods concern large moving objects but they only detect them and do not correct distortion \cite{Gepshtein:restoration:2004, Fishbain:realtime:2007, Halder:geometric:2015, Deshmukh:embeded:2016}.

Our previous work, `CLEAR' (Complex waveLEt fusion for Atmospheric tuRbulence), employs an image fusion technique in the wavelet domain, which has proved its capability in a variety of turbulent atmospheric environments  \cite{Anantrasirichai:mitigating:2012, Anantrasirichai:Atmospheric:2013}. The method exploits  existing information that is already possessed in the sequence. Moreover, denoising, sharpening and contrast enhancement, if necessary, can be performed simultaneously in the wavelet domain \cite{Loza:automatic:2013}. 

In this paper, we further develop the wavelet-based fusion method for distorted sequences that contain moving objects (CLEAR2). We apply motion-based tracking via a Kalman filter and model the background with Gaussian mixture models (GMM). These deal effectively with the uncertainty inherent in noisy data. The measurement of the object location is already integrated with a non-rigid registration process \cite{Chen:registation:2011}, employed to shift turbulence displacement. However, sometimes objects move faster than the ability of non-rigid registration. Hence, we also provide an object warping process for motion compensation. 
The sequence is restored in a recursive manner, widely used to minimise buffer size requirements, computational complexity \cite{Deshmukh:embeded:2016} and the propagation of uncertainty \cite{Anantrasirichai:terrain:2015}. 

The remainder of this paper is organised as follows.  Section \ref{sec:existingwork} describes related work and Section \ref{sec:proposedscheme} presents the proposed method. The performance of the method is evaluated in Section \ref{sec:results}. Finally, Section \ref{sec:conclude} presents the conclusions.

\begin{figure}[t!]
	\centering
  		\includegraphics[width=\columnwidth]{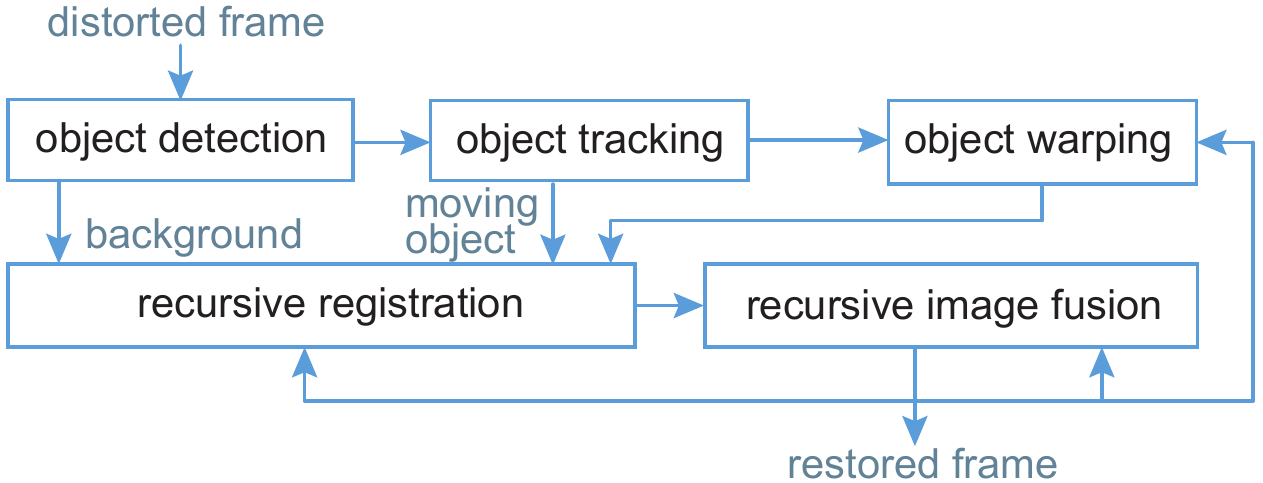}
	\caption{Diagram of the proposed method.}
	\label{fig:recursivediagram}
\end{figure} 

\section{Related work}
\label{sec:existingwork}

The existing methods generally employ an image registration technique with deformation estimation \cite{Zhu:Removing:2013,Anantrasirichai:Atmospheric:2013}. This process attempts to align objects temporally to solve for small movements of the camera and temporal variations due to atmospheric refraction. Then, these registered images are averaged or image fusion is employed in order to combine several aligned images \cite{He:atmospheric:2016}. A deblurring process is finally applied to this combined image  \cite{Zhu:Removing:2013}. Most methods in the literature however have been proposed for static scenes. Reducing atmospheric turbulence effects in a video requires an additional motion model when the objects in the scene are themselves moving. 

To detect moving objects in the turbulent atmospheric medium, most methods reconstruct the static background first and employ thresholding techniques using motion vectors and/or intensity \cite{Gepshtein:restoration:2004, Fishbain:realtime:2007}.   
Halder et. al. proposed an iterative approach to remove turbulent motion and the moving object is masked using simple thresholds on both motion and intensity \cite{Halder:geometric:2015}. Unfortunately,  the method did not remove the distortion around the moving object.
A low-rank matrix approach decomposing the distorted image into background, turbulence, and moving object is presented in \cite{Oreifej:Simultaneous:2013}. An adaptive threshold technique applied to the background model using a temporal median filter in \cite{Chen:detecting:2014}.

Three methods in the literature were introduced to mitigate turbulent distortion for large moving objects. The moving objects are detected using block matching techniques in \cite{Huebner:Software:2011, Foi:methods:2015}. These are employed to separate the two types of motion with an assumption that the object movement is larger than the turbulent flow. The compensated moving areas are aligned in the 3D volume and the turbulent distortion on these areas is suppressed in the same way as the static background areas.  In \cite{Foi:methods:2015}, the true motion is also estimated by smoothing the motion trajectories to remove small random movement caused by turbulence across a fixed number of successive frames. The authors in \cite{Kelmelis:practical:2017} developed `\textit{dynamic local averaging}', which determines the number of frames to employ for averaging, to avoid any unwanted effects. This method however may not mitigate the distortion on the moving objects as mostly only one frame is employed.

\section{Proposed Scheme}
\label{sec:proposedscheme}

The proposed method is depicted in Fig. \ref{fig:recursivediagram} and the functionality of each block is explained below.

\vspace{3mm}
\noindent \textbf{3.1. Object detection and tracking}
\vspace{2mm}

\noindent For a new frame $X_t$ at time $t$,  the process starts with foreground (FG) and background (BG) separation. A background subtraction technique based on a Gaussian mixture model (GMM)  is employed \cite{Stauffer:adative:1999} . We improve the model by including probability density functions (pdf) of the motion estimated in the non-rigid registration process (Section 3.2). The weight, mean and variance of each distribution is updated in recursive manner. The BG mask $M^B_t$ represents the region where the summation of distributions is larger than a threshold. Assuming the area of the BG is always larger than that of the FG, we set the threshold using the median value of all distributions.
To track an object $k$ from a total of $K$ moving objects, the motion of the centroid of each FG mask $M^{F,k}_t$ is estimated using a Kalman filter. Briefly, Kalman filtering employs Bayesian inference and a joint probability distribution over the measured variables for each frame to estimate the locations of the observed objects.  In this paper, the objects are assumed to be moving with constant velocity. For nonlinear systems, the extended Kalman filter (EKE) or unscented Kalman filter (UKF)  can be employed \cite{Wan:unscented:2000}.

\vspace{3mm}
\noindent \textbf{3.2. Motion estimation through non-rigid registration}
\vspace{2mm}

\noindent Registration of non-rigid bodies using the phase-shift properties of Dual-Tree Complex Wavelet Transform (DT-CWT) coefficients is employed, similar to our previous work \cite{Anantrasirichai:Atmospheric:2013}. Motion estimation is performed by firstly using coarser level complex coefficients to determine large motion components and then by employing finer level coefficients to refine the motion field. It should be noted that no more complexity is added to the framework for motion estimation used for object detection in Section 3.1. 

\vspace{3mm}
\noindent \textbf{3.3. Object warping}
\vspace{2mm}

\noindent We know translation parameters for each moving object from Section 3.1.  However, sometimes their movements are not simply pure translation and hence the motion compensation is not  good enough to produce reasonable results using non-rigid registration.  Therefore, we also introduce a warping process using multi-scale gradient matching \cite{Farid:video:2007}. This function is activated when the error from Section 3.2 exceeds a threshold. The $2\times 2$ affine matrix $A_{t,t-1}$  and the $2 \times 1$ translation vector $T_{t,t-1}$ linked between two consecutive frames $t$ and $t-1$ are computed using the highpass coefficients of the DT-CWT  extracted within the $M^{F,k}_{t}$  and $M^{F,k}_{t-1}$, respectively.
The moving object area $O^k_t$ is constructed in recursive manner as shown in Eq. \ref{eqn:O_tn}, where $\alpha$ is learning rate.
\begin{equation}
\label{eqn:O_tn}
		O^k_t = M^{F,k}_t \left( (1-\alpha) (A^k_{t,t-1} O^k_{t-1} + T^k_{t,t-1}) + \alpha  X_t \right)
\end{equation}

\vspace{5mm}
\noindent \textbf{3.4. Recursive registration}
\vspace{1mm}

\noindent A recursive strategy is proposed for updating the reference $R_t$ at time $t$. Subsequently, the current input frame  $X_t$ is non-rigidly registered to $R_t$, which happens only once per frame thereby significantly reducing the workload. The reference frame can simply be updated by adding a new frame in and subtracting the oldest frame out of the summation, but this system can develop error build-up over long time periods. Therefore, we create $R_t$ with exponentially decaying weights as shown in Eq. \ref{eqn:recursiveRef}, where $\alpha$ is the same parameter in Eq. \ref{eqn:O_tn}. $N_b$ and $N_f$ are the numbers of previous frames to restore the BG and FG, respectively. Generally, $N_b > N_f$. We set $\alpha = 1/(N_b+1)$, which is approximately equivalent to averaging the last $N_b+1$ frames.  The current frame $X_t$ is registered to $R_t$ using the method in Section 3.2.
\begin{equation}
\label{eqn:recursiveRef}
		R_t = M^B_t [(1-\alpha) R_{t-1} + \alpha X_t] + (1-M^B_t) \sum^K_k  O^k_{t}
\end{equation}

\noindent \textbf{3.5. Recursive image fusion}
\vspace{2mm}

\noindent We denote $\{a^{\theta}, d^{\theta,l}\}_L = \Psi(X,L)$, where DT-CWT $\Psi (\bullet)$ decomposes an image $X$ into lowpass subbands $a^{\theta}$ and highpass subbands $d^{\theta,l}$, $\theta \in \{\pm 15^{\circ}, \pm 45^{\circ}, \pm 75^{\circ}\}$,  $l \in \{1, ..., L\}$, and $L$ is the total decomposition level. At time $t$, the distorted frame $X_t$ is registered to $R_t$, resulting $X^R_t$. Then, we have $\{v^{\theta}_{R_t}, w^{\theta,l}_{R_t}\}_L  = \Psi(R_t,L)$ and $\{v^{\theta}_t, w^{\theta,l}_t\}_L  = \Psi(X^R_t,L)$.
The mask $m^{l}$ is the resized version of ($M^B_t M^B_{t-1} > 0$) with the same size of $w^{\theta,l}$.

In the recursive image fusion,  $a_t^{\theta}$ is constructed as described in Eq. \ref{eqn:recursivelow}, and $d_t^{\theta,l}$  is constructed  following Eq.  \ref{eqn:recursivehighang} and Eq.  \ref{eqn:recursivehighabs}. 
The angle of  $d_t^{\theta,l}$ is also merged using  $\alpha$ to give exponentially decaying weight to those of previous frames.
However, applying this idea to the absolute value of the coefficients will not be able to produce a sharp fused frame, since the high frequencies present in previous frames are diminished. Therefore we propose using a weight  $\beta$ with binary mask $Q_t^{\theta,l}$, which is set to 1 if the current wavelet magnitude is smaller than the median value of all coefficients in the same subband and level ($\text{med}(z)$ is the median value of data $z$ in Eq. \ref{eqn:recursivehighabs}). This ensures that strong structures, e.g. corners and lines, are sharp and the accumulated high frequencies presented in the homogeneous areas are suppressed to prevent undesired artefacts. 
Finally, the restored frame is produced as $Y_t =\Psi^{-1}_L (\{a^{\theta}_t, d^{\theta,l}_t\})$, where $\Psi^{-1}_L (\bullet) $ is an inverse DT-CWT. 
\begin{equation}
\label{eqn:recursivelow}
		a_t^{\theta} = m^{L} [(1-\alpha) a^{\theta}_{t-1} + \alpha v_t^{\theta}] + (1-m^{L})  v^{\theta}_{R_t}
\end{equation}
\small
\begin{equation}
\label{eqn:recursivehighang}
		\angle d_t^{\theta,l}   = m^{l}  \frac{(1-\alpha) d_{t-1}^{\theta,l} + \alpha w_t^{\theta,l}}{|(1-\alpha) d_{t-1}^{\theta,l} 	+ \alpha w_t^{\theta,l}|}  + (1-m^l) \angle w^{\theta,l}_{R_t}
\end{equation}
\begin{equation}
\label{eqn:recursivehighabs}
\small
\begin{gathered}
		|d_t^{\theta,l} | = \max \left( \beta m^{l}  |d_{t-1}^{\theta,l} |, |w_t^{\theta,l}| \right)   \\
		\beta    = 1 - \alpha Q_t^{\theta,l}, \; \; Q_t^{\theta,l} = \frac{ |w_t^{\theta,l} |}{  \text{med}( |w_t^{\theta,l}| ) }<1
\end{gathered}
\end{equation}

\normalsize
\section{Results and discussion}
\label{sec:results}

The proposed method was evaluated with seven sequences, namely i) Car in Dubai ii) Two people at 1.5km, iii) People with tools, iv) Van driving in circles at 0.75km, v) Dodge in heat wave, vi) Train in strong heat, vii) Plane in airport. The first sequence was captured by our team (VI-Lab, University of Bristol). Sequences 2-4 were provided by DSTL, and the rest were acquired from YouTube. All these sequences are available on the VI-Lab website\footnote{eis.bris.ac.uk/$\sim$eexna/download.html}.  If not stated, the results were restored with $L=4$, $N_b=50$ and $N_f=5$.

\begin{figure*}[t!]
  		\includegraphics[width=\textwidth]{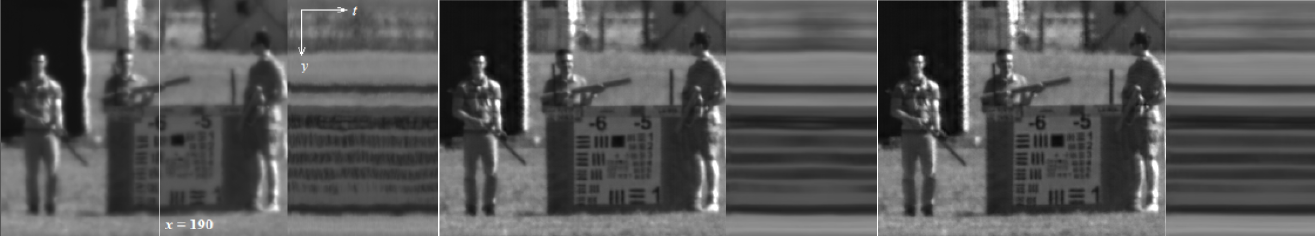}
	\caption{Results of `\textit{People with tools}' sequence (Left), using a registration method with a sliding window of $N_b=50$ (Middle) and a recursive technique (Right). In each group, the left image shows the frame at $t=150$ and the right image shows the $yt$ planes at column $x=190$ of this frame in the first 200 frames.}
	\label{fig:registercompare}
\end{figure*}

Firstly we examined the performance of the proposed recursive technique (CLEAR2) compared to transitional temporal sliding window used in CLEAR. The subjective results are shown in  Fig. \ref{fig:registercompare}. Both methods produced videos that are clearly more stable than the original. The difference between the two registered results was hardly noticeable, whilst the processing time was reduced by 20-fold in this test.

\begin{figure}[t!]
	\centering
  		\includegraphics[width=\columnwidth]{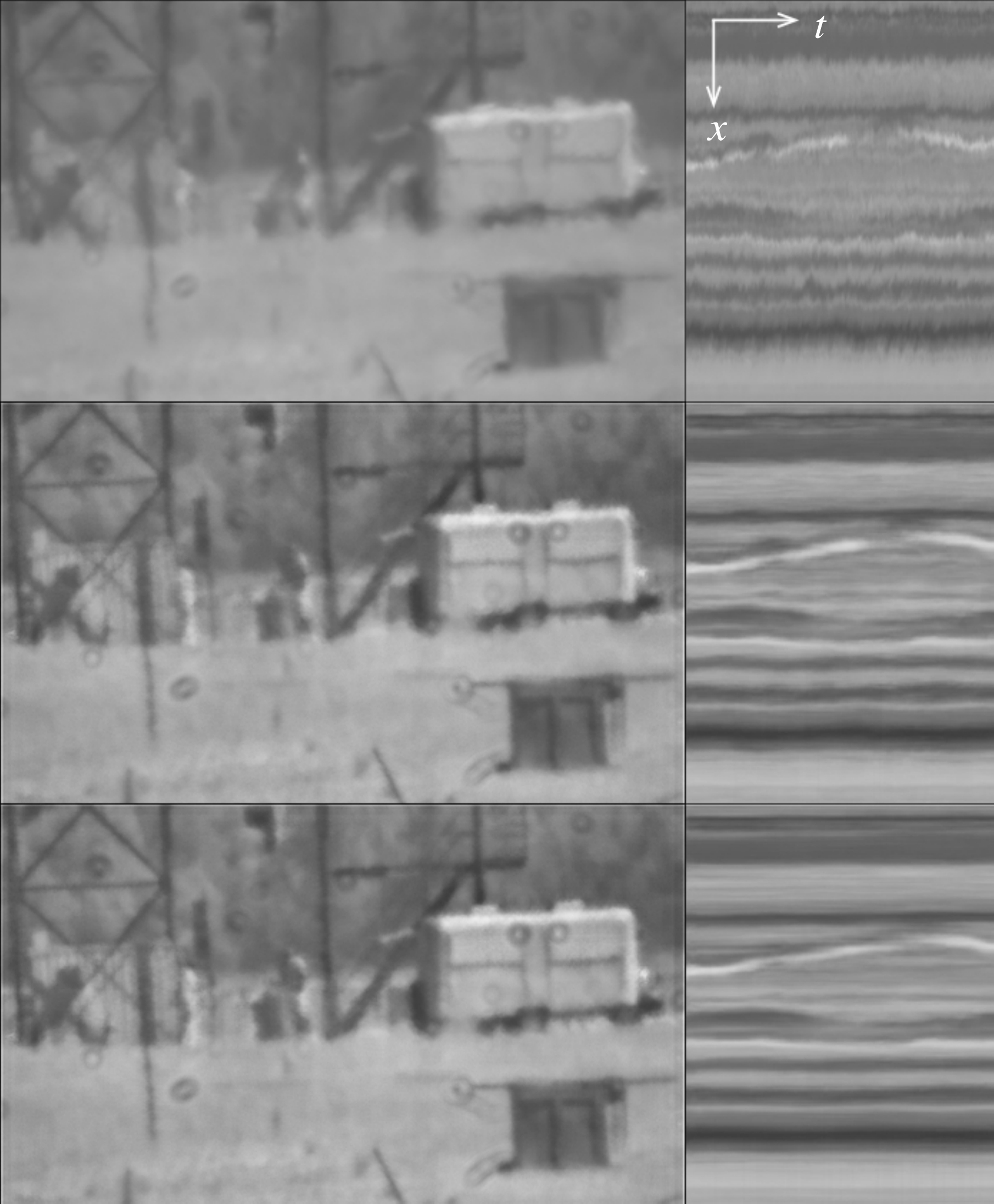}
	\caption{Results of  `\textit{Two people at 1.5km}' sequence (top) enhanced with (middle) $N_b=20$ and (bottom) $N_b=50$. Left column: frame $t=100$. Right column: the $xt$ plane at $y=180, x \in [1,440], t \in [730,100]$.}
	\label{fig:compare_Nb}
\end{figure}

Fig. \ref{fig:compare_Nb} shows the effects of different $N_b$. The result in the middle row of this figure was generated using smaller $N_b$ than that in the bottom row. The cropped $xt$ plane at $y=180$ of these results are presented in the right column. It is clear that the larger $N_b$ offers higher contrast and smoother in temporal direction which is good for reconstructing static background. However, the unsuccessful warped object areas from the earlier frames may be accumulated and present as, for example, unclear edges in the area between the left man and the pole in the picture on the bottom row.

\begin{figure}[t!]
	\centering
  		\includegraphics[width=\columnwidth]{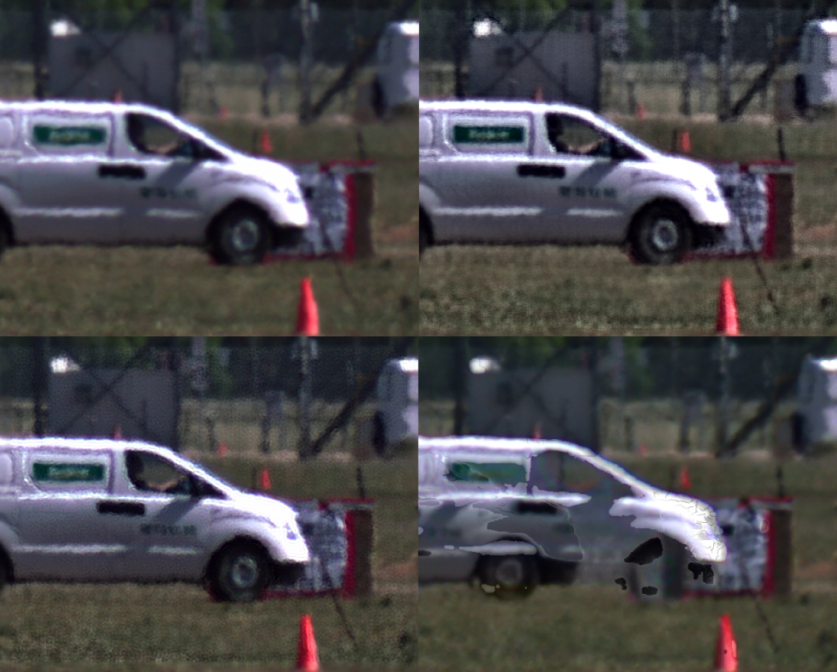}
	\makeatother
	\caption{Frame $t=19$ of `\textit{Van driving in circles at 0.75km}' sequence. (top-left) distorted frame. (top-right) CLEAR2 (coarse registration). (bottom-left) DLA adapted from \cite{Kelmelis:practical:2017}. (bottom-right) EVS \cite{Deshmukh:embeded:2016}.}
	\label{fig:compare_van}
\end{figure}

We compared our method with two existing methods: i) Embedded vision system (EVS) \cite{Deshmukh:embeded:2016} and ii) Dynamic local averaging (DLA) adapted following \cite{Kelmelis:practical:2017}. Fig. \ref{fig:compare_van} shows the results of `\textit{Van driving in circles at 0.75km}' sequence and 
Table \ref{tab:performance} show the average computational time of all seven test seqeunces. All methods were implemented in Matlab, CPU i7-3770S, 16GB RAM.
The quality of the restored video from CLEAR2 is better, particularly when compared with EVS -- obviously simple intensity and colour thresholding technique does not work for large moving objects. Our result is the sharpest and the ripple distortion is mitigated most. Some areas in the result of DLA still contain atmospheric distortion, whilst some areas appear to be over-sharpened. The performance of CLEAR2 in terms of computational time is significantly improved compared to the previous CLEAR as shown in Table \ref{tab:performance}. However, CLEAR2 is slower than EVS and DLA because of the non-rigid registration process. We tested CLEAR2 with non-rigid registration operating only on the coarse level (level $L$ of DT-CWT), which reduced the computational time by half, whilst preserving the majority of the turbulence mitigation.
Note that implementation in Field-programmable gate array (FPGA) or GPU should speed up the process, e.g. 15 times in \cite{Deshmukh:embeded:2016} and 45 times in \cite{Kelmelis:practical:2017}, respectively.

\begin{table}[t!]
	\centering
	\caption{Computational performance comparison (sec/frame)}
	\small
		\begin{tabular}{c|ccc}
		\hline
			\multirow{2}{*}{method} & \multicolumn{3}{c}{resolutions} \\ \cline{2-4}
			& 224$\times$320 & 576$\times$704  & 720$\times$1280\\
			\hline
			EVS \cite{Deshmukh:embeded:2016} & 0.25 & 1.59 & 3.18 \\
			DLA adapted from \cite{Kelmelis:practical:2017} & 1.58 & 6.25 & 10.67\\
			\hline
				CLEAR (5 references) & 5.32 & 18.24 & 25.97\\
				CLEAR (20 references) & 16.32 & 55.39 & 78.23\\
				CLEAR2 full registration  & 1.64 & 8.69  & 13.05\\
				CLEAR2 coarse registration &  0.65 & 3.87 & 7.01 \\
				\hline 
		\end{tabular}
	\label{tab:performance}
\end{table}

\section{Conclusions}
\label{sec:conclude}

This paper has introduced a new method for mitigating atmospheric distortion in long range surveillance imaging. The improvement of visibility of moving objects in observed sequences is achieved using recursive image fusion in the DT-CWT domain. The moving objects are detected and tracked using modified GMM and Kalman filtering. Both background and moving objects are restored by adding the current frame to the previous result with exponentially decaying weight. With recursive registration and fusion, our CLEAR2 technique improves computational performance over the previous CLEAR. 
We also introduce a coarse-registration option which achieves comparable speed to competing methods with significantly better subjective quality.

\newpage
\balance

\end{document}